\documentclass[runningheads]{llncs}
\usepackage{graphicx}
\usepackage{subcaption}
\usepackage{multirow}
\usepackage{booktabs} 

\usepackage{tikz}
\usepackage{comment}
\usepackage{amsmath,amssymb} 
\usepackage{color}
\usepackage{bm} 
\usepackage[ruled,vlined]{algorithm2e}

\usepackage{caption}

\begin{document}
\pagestyle{headings}
\mainmatter

\title{Large Scale Neural Architecture Search with Polyharmonic Splines} 



\titlerunning{Large Scale Neural Architecture Search on Imbalanced Data with Polyharmonic Splines}
%
\author{Ulrich Finkler\inst{1} \and
Michele Merler\inst{1} \and
Rameswar Panda\inst{1} \and
Mayoore S. Jaiswal\inst{2} \and
Hui Wu\inst{1} \and
Kandan Ramakrishnan\inst{3} \and
Chun-Fu Chen\inst{1} \and
Minsik Cho\inst{4} \and
David Kung\inst{1} \and
Rogerio Feris\inst{1} \and
Bishwaranjan Bhattacharjee\inst{1}}
\institute{$^{1}$ IBM Research, $^{2}$ NVIDIA, $^{3}$ Baylor College, $^{4}$ Apple}
\maketitle
\setcounter{footnote}{0}
\renewcommand{\thefootnote}{\roman{footnote}}
\begin{abstract}
Neural Architecture Search (NAS) is a powerful tool to automatically design deep neural networks for many tasks, including image classification. Due to the significant computational burden of the search phase, most NAS methods have focused so far on small, balanced datasets. All attempts at conducting NAS at large scale have employed small proxy sets, and then transferred the learned architectures to larger datasets by replicating or stacking the searched cells. We propose a NAS method based on polyharmonic splines that can perform search directly on large scale, imbalanced target datasets. We demonstrate the effectiveness of our method on the ImageNet22K benchmark\footnote{ImageNet was used for research purposes only to allow benchmarking against prior results from others on this data set. The trained models in this work are not used for commercial purposes.}~\cite{imagenet_cvpr09}, which contains $14$ million images distributed in a highly imbalanced manner over $21,841$ categories. By exploring the search space of the ResNet~\cite{resnet} and Big-Little Net ResNext~\cite{blresnext} architectures directly on ImageNet22K, our polyharmonic splines NAS method designed a model which achieved a top-1 accuracy of $40.03\%$ on ImageNet22K, an absolute improvement of $3.13\%$ over the state of the art with similar global batch size~\cite{intel_blog_2017}.

\end{abstract}

\section{Introduction} \label{sec:intro}
Designing Deep Neural Networks is a challenging task and requires a Subject Matter Expert (SME). 
One way of reducing the design burden and still be able to obtain a custom designed architecture for a given training problem is to use Neural Architecture Search (NAS)~\cite{survey_JMLR2019,NAS_RL_ICLR2017}.
Most NAS methods have been tried on small scale datasets like Cifar10-100~\cite{Krizhevsky09learningmultiple} and then subsequently architectures have been transferred to larger datasets like ImageNet1K~\cite{Borsos2019TransferNK,lu2020neural,wistuba2019xfernas}. This is mainly due to the heavy computational burden of the search phase, which despite recent progress~\cite{liu2018darts,Zela2020NAS-Bench-1Shot1}, remains mostly intractable at large scale. No previous published work has explored NAS on large scale datasets like ImageNet22K directly, and very few have even attempted it on ImageNet1K itself.

Additionally, most NAS methods have been tested on datasets with uniform distribution in terms of number of images per class. For example Cifar100 has $100$ classes containing $600$ images each, and Imagenet1K has 1k classes with 1k images each. While the design of such uniform distributions facilitate learning, they are not reflective of real world conditions, where skew in terms of instances per label/class is very much evident. 

In this paper we describe a novel method for NAS that can conduct search directly on large-scale skewed datasets using Polyharmonic Splines. First, we describe in detail how specific operations within a deep neural network (convolution, ReLu, average pooling, batchnorm) are well suited to be approximated by a spline. Given a search space consisting of a set of $d$ operations to optimize over, the search phase of our proposed NAS method based on polyharmonic splines requires only an initial set of $2d + 3$ points, followed by few additional points to complete its the optimization. This means that the number of evaluations required in the search phase depends only on the number of operations under search, not on the number of possible values for each of those operations. This drastically reduces the computational cost of the search phase of our NAS approach, allowing it to be performed directly on large scale target datasets.

We demonstrate the effectiveness of our method on the ImageNet22K benchmark of  $14$ million images over $21,841$ labels. The dataset has high skew in terms of images per label and distribution across semantic categories, as shown in Figure~\ref{fig:22kimgs}. By exploring the search space of the ResNet~\cite{resnet}, a Big-Little Net~\cite{blresnext}, and ResNext~\cite{xie2017aggregated} architectures, the method designed a model which achieved a top-1 accuracy of $40.03\%$ on ImageNet22K. This significantly improves the best published performance~\cite{intel_blog_2017} of $36.7\%$ on such large-scale, imbalanced dataset 

The remainder of the paper is organized as follows. After Section~\ref{sec:related}, which covers related work, we describe in Section~\ref{sec:basis} the basic components of deep neural networks which are adopted as dimensions for our type of NAS and explain why the functions on such variables are well suited for interpolation. We describe the application of polyharmonic splines and the necessary conditions for the existence of a solution in Section~\ref{sec:NASspline}, before discussing experimental results on ImageNet22K in Section~\ref{sec:experiments}. Finally we draw conclusions in Section~\ref{sec:conclusions}.

\section{Related Work} \label{sec:related}

\textbf{NAS.}
Neural Architecture Search (NAS), which aims at automatic design of deep learning networks for various applications spanning from image classification \cite{progressiveNAS_ECCV2018,MnasNet_CVPR2019,wu2018fbnet,randomly_wired_ICCV2019} to NLP \cite{liu2018darts,pmlr-v80-pham18a,NAS_RL_ICLR2017}, from object detection \cite{DetNAS_Neurips2019,NASFPN_CVPR2019,wang2019nasfcos} to semantic segmentation \cite{liu2019autodeeplab} and control tasks \cite{weight_agnostic_neurips2019} has attracted intense attention in recent years.  
A number of NAS strategies have been proposed, including evolutionary methods \cite{baker2018accelerating,liu2018hierarchical,regularized_AAAI2019,EcoNAS_CVPR2020}, reinforcement learning \cite{progressiveNAS_ECCV2018,pmlr-v80-pham18a,ZhongYWSL18,NAS_RL_ICLR2017}, and gradient-based methods \cite{DATA_neurips2019,liu2018darts,nayman2019xnas,xie2019snas,Zela2020Understanding}. 
Efficiency on specific platforms has also been a very active area of research within the NAS umbrella, with the development of search strategies that optimize not only accuracy but also latency \cite{cai2019proxylessnas,howard2019searching,MnasNet_CVPR2019,tan2019efficientnet,wu2018fbnet}. Methods based on micro-search of primary building cells \cite{ZhongYWSL18,ZophVSL17}, and parameter sharing between child models \cite{pmlr-v80-bender18a,brock2018smash,liu2018darts,pmlr-v80-pham18a,Zela2020NAS-Bench-1Shot1} have also been recently proposed. Another widely used approach to address efficiency in NAS is to search for an architectural building block on a small dataset (e.g., CIFAR10) and then transfer the block to the larger target dataset by replicating and stacking it multiple times in order to increase network capacity according to the scale of the dataset. \\

\noindent\textbf{Bayesian Methods.} Our work is also related to Bayesian optimization which often use an acquisition function to obtain suitable candidates, has been widely used in NAS~\cite{domhan2015speeding,mendoza2016towards,kandasamy2018neural,cao2019learnable}. The acquisition function measures the utility by accounting for both, the predicted response and the uncertainty in the prediction. While Bayesian optimization methods usually consumes more computation to determine future
points than other methods, this pays dividends when the evaluations are very expensive. Many different strategies have been studied including classic Gaussian process-based optimization~\cite{swersky2014raiders,kandasamy2018neural}, tree-based models~\cite{bergstra2013making}, random forests~\cite{falkner2018bohb} to effectively optimize both neural network architectures and their hyperparameters. Several works also predict the validation accuracy~\cite{zhang2018graph}, or the curve of validation accuracy with respect to training time~\cite{baker2018accelerating}.    

\section{Optimization Basis} \label{sec:basis}

Neural Architecture Search in its essence optimizes a function $y=f(\vec{x})$, $x$ residing in a multidimensional space of network and training parameters and $y$ being a metric for the quality of the network, for example the top-1 accuracy on a validation data set.
The parameters captured in $\vec{x}$ may be discrete. $f$ might not be differentiable in terms of the parameters $x_i$.
Importantly, $f$ is generally very expensive to evaluate. For a given parameter set $x$, the network has to be trained and evaluated on the full training and validation sets.

The interpolation quality of methods to approximate scalar functions of multidimensional arguments based on a limited set of support points depend on the properties of the interpolated function.
Hence it is important to understand the properties of $f$ and if it is likely that there exists a well behaved interpolating function $F:R \rightarrow R^N$ through the potentially discrete support points.

This section discusses the algebraic structure of popular residual neural networks and establishes that the resulting $f(\vec{x})$ is with high probability well suited for interpolation with a polyharmonic spline for parameters that have a sufficiently large number of possible values. For parameters with only a few choices, using separate splines for the discrete cases can be more effective. Furthermore, the algebraic analysis provides insights into the sensitivity of global network parameters and hence search spaces.

\subsection{Neural Network Forward Pass as Piecewise Linear Function}

The forward pass through a typical  neural network consisting 
of convolutions, fully connected layers, ReLUs, batch-norms and average-pooling 
can be interpreted as a piecewise linear function that effectively transforms an input, 
for example an image as a set of $224\times224$ values across three channels, 
into $d_f$ values in a {\em feature space}, which forms the input 
to a final linear classifier. 

\subsubsection{Fully Connected Layers}

A single neuron with $N$ inputs performs the operation $y = \phi \left( \sum_{i=1}^{N} w_i x_i + b \right)$
with an activation function $\phi$. In the following we discuss the case $\phi(x) 
= max(0,x)$, i.e. ReLU. 
 
The condition $0 = \sum_{i=1}^{N} w_i x_i + b$ defines a hyperplane in the $N$ dimensional 
input space of $\vec{x}$ such that $y>0$ for locations above the hyperplane and $y=0$ 
for locations on or below the hyperplane. The output of the neuron without activation is in fact 
the distance of $\vec{x}$ from the hyperplane. 
Multiple neurons with inputs $\vec{x}$ from the same input space define a set of hyperplanes 
that partition the input space into a set different regions. In each region, $\vec{y}$ is determined by 
a set of linear equations. 

For example in a two-dimensional input space, 2 neurons $y^{[1]}$ and $y^{[2]}$ with linearly 
independent weight vectors and ReLU activation create four regions
\begin{eqnarray*}
  \vec{y} = (\sum_{i=1}^{N} w_i^{[1]} x_i + b^{[1]} ,  \sum_{i=1}^{N} w_i^{[2]} x_i + b^{[2]})  \quad;\quad \vec{y} = (0,0)\\
  \vec{y} = (\sum_{i=1}^{N} w_i^{[1]} x_i + b^{[1]} , 0 ) \quad;\quad 
  \vec{y} = (0 , \sum_{i=1}^{N} w_i^{[2]} x_i + b^{[2]}) 
\end{eqnarray*}

A second layer of similar structure, a linear operation followed by ReLU activation, potentially partitions each of the segments of the input space again. 
In the following we discuss how other layers, namely \textit{convolution}, \textit{batchnorm} and \textit{average pooling}, are operations of analogous structure and hence the entire set of 
operations leading to the final linear classifier is a piecewise linear function that maps hyper-volumes 
in the input space into the feature space. For a trained neural network, this piecewise linear 
function is optimized to create linearly separable clusters of mapped training points in the feature space. 

\subsubsection{Convolution}

Convolution in a neural network is
\[
  y_{ij}^{(k)} = \sum_{a,b,c} W_{abc}^{(k)} x_{(i+a)(j+b)c}
\]
wherein $c$ identifies the input channel, e.g. color in an rgb image or filter from a prior convolution. 
The tuple $(i,j)$ is the location within the input at which a {\em patch} of the size of a filter is scanned. 
$k$ identifies the filter, i.e. the output channel of the convolution. 
$a$ and $b$ iterate through the positions within the input patch and filter. Hence, each tensor $\bar{W}$ is of 
size $N_A*N_B*N_C$, the filter width, height and number of input channels. 

We can apply an index transformation $\gamma = a*N_B*N_c + b*N_C+ c$ which transforms $W_{abc}^{(k)}$ 
into $W_{k\gamma}$. For a fixed position $(i,j)$ of the convolution we can drop the indices $i$ and $j$ to obtain $
  y_k = \sum W_{k\gamma} x_\gamma $.
The vectors $\vec{x}$ are points in a vector space of dimension $N_A*N_B*N_C$. I.e., on one input patch
 a convolution has the same algebraic structure as a neuron. Average-pooling is a convolution with 
 a filter whose elements are all identical. 

Note that these operations can be expressed in the form of a fully connected layer, with a set of neurons that correspond 
to the filter channels that is present multiple times with different subsets of the input weights set to 
zero to express the selection of patches in the input. 

\subsubsection{Batch Norm} 

On first glance, a batch norm layer does not look like a linear function. But a closer look at the details 
reveals that the running means and averages are treated as constants in backpropagation, they do not 
have gradients. The batch norm (mean and variance across the batch, width and height for a channel) 
is an estimate for the constant that is used during inference on the trained model, which is in 
itself an estimate for the mean and variance across the entire training set. As the weights change 
during training, the estimates for running means and variances follow.  

Pytorch uses Bessel's correction $s = \sigma \frac{n}{n-1}$, an unbiased estimate of the population variance $\sigma = \frac{1}{n} \sum_i (x_i - \mu)^2 $ from a finite sample for {\em BatchNorm2d}.

With  $x_{ijkl}$ as an input tensor of form NCWH, $n = \sum_{ikl} 1$ being the number of elements in a 'channel slice' and $t$ as the iteration number, 
mean and unbiased variance for a 'channel slice' are
\begin{eqnarray*}
    \mu_j(t) &=& \frac{1}{n} h_j(t) \quad   with \quad h_j(t) = \sum_{ikl} x_{ijkl}(t) \\
    \psi_j(t) &=& \frac{1}{n-1} g_j(t) \quad with \quad  g_j(t) = \sum_{ikl} ( x_{ijkl}(t) - \mu_j(t))^2
\end{eqnarray*}
With running mean and variance
\begin{eqnarray*}
M_j(t) &=& (1-m)M_j(t-1)+m\mu_j(t)   \quad with \quad M_j(0) = 0 \\
S_j(t) &=& (1-m)S_j(t-1)+m\psi_j(t) \quad with \quad S_j(0) = 1
\end{eqnarray*}
batch norm in training ($B$) and eval ($B'$) mode result in
\[
  B(x_{ijkl}(t)) = \frac{ x_{ijkl}(t)-\mu_j(t) }{\sqrt{\sigma_j(t)+\epsilon}} \quad\quad  B'(x_{ijkl}(t)) = \frac{ x_{ijkl}(t)-M_j(T) }{\sqrt{S_j(T)+\epsilon}}
\]
Hence, with $M$ and $S$ as constants, batch norm maps a linear transformation.

\subsection{Base Transformations, Projections and Convolutional Networks} 

  The rank-factorization $\bar{A} = \bar{C}\bar{B} $ of a $m \times n$ matrix $\bar{A}$ of rank $r$ 
expresses $\bar{A}$ as a product of a $m \times r$ full column rank matrix $\bar{C}$ and a 
$r \times n$ full row rank matrix $\bar{B}$. 
For a square matrix of full rank the equation $ y_k  = \sum_{i=1}^{N} w_{ki} x_i $ 
is a transformation of vector $x_i$ from one base of $R^N$ to another.

Hence, conceptually a $m \times n$ matrix of rank $r$ maps a vector from 
a space with $m$ dimensions into a subspace of $r$ dimensions and from there 
into a space with $n$ dimensions. If $r<n$, then we can express all of the 
$n$ dimensional vectors in terms of an $r$ dimensional basis. Thus, a trained 
network whose matrices are not close to full rank is inefficient. 

Furthermore, we can interpret a convolution and subsequent activation on a single 'patch' 
as a transformation from a 
$m = w*h*c_{in}$ dimensional space to a $n=c_{out}$ dimensional space, $w$ and $h$ being the 
width and height of the filters and $c_{in}$ and $c_{out}$ being the numbers of the input 
channels and output channels, respectively. 

Assuming that the combined convolution matrix $\bar{W}_{k\gamma}$ has close to full rank in an efficient 
network, the activations resulting from a convolution can be interpreted as the coefficients 
in an expression that approximates the input tensor in terms of a set of functions defined 
by the filter tensors, a form of compression. 

The 'approximation' has to be of limited loss, since the correlation between classification results 
and inputs has to be preserved. In some sense training optimizes the approximation such that it 
aides in transforming inputs into linearly separable points in the {\em feature space} and at the same time 
it allows to approximate with limited loss of information across all its inputs.


  Based on above observations, the essential part of the functionality of the 
{\em convolutional cone} leading to a linear classifier in a neural network is provided by 
mapping points from one space to another 
to create a piecewise linear function whose output in the feature space is highly linearly separable 
with respect to the number of hyperplanes in the final classification layer. What primarily matters 
is how many linear pieces the network provides relative to the number of hyperplanes, i.e. classes, 
in the final classifier. 

A second aspect is trainability, i.e. how easy a network converges, 
which suggests to focus on residual networks with batchnorm. While the specific structure of a network 
should have relatively little influence on final accuracy, it may have a significant impact 
on the network size that is required to generate a competitive piecewise linear approximation. 
Which lead us to focus architecture search on different families of residual networks,
specifically ResNet~\cite{resnet}, ResNeXt~\cite{resnext} and BLResNeXt~\cite{blresnext}.
Typical residual networks have 4 {\em groups} of residual blocks with different numbers of layers. 
For example ResNet50 has groups of input width 64, 128, 256, 512 channels with an expansion factor of 4 and 
depths of 3, 4, 6, 3. ResNet18 on the other hand has the same widths, but an expansion factor of 1 and 
depths of 2, 2, 2, 2. 

The number of classes, i.e. the number of hyperplanes in the feature space, provides guidance on the optimal 
number of dimensions for the feature space. In an $N$ dimensional space, we can place $N$ linearly independent 
hyperplanes such that all volumes bounded by hyperplanes are infinite. If we place more hyperplanes, some 
volumes have to be limited to a constant that depends on the placement of the hyperplanes. 
Hence, a dimesion of the feature space that is larger than the number of classes should be beneficial. 
Note that for ResNet50 the dimension of the feature space is 2048 and for ResNet18 it is 512. 

If the leading layers 'choke' the dimensions of the vector spaces leading to the feature space too much, 
the analysis suggests that this will negatively influence the accuracy of the network. 
For example the initial convolution in resnet 18 has an input space of $7\times7\times3=147$ dimensions and 
maps this to 64 dimensions. A $3\times3$ convolution with 64 channels to 128 channels maps 576 dimensions to 
128, i.e. creates a significant reduction. Furthermore, the number of activations drops by a factor of 4 
through every group. 

As an experimental test we designed a simple neural network r18U based on resnet18 whose behavior should be 
closer to resnet50 for Imagenet1k by adjusting only the block widths. 
We eliminated the {\em max-pool} layer and compensated for its reduction by increasing the stride of the 
initial convolution to $2$ based on the hypothesis 
that the non-linear layer is not essential for the functionality of the neural network. Indeed, this did not 
negatively impact final accuracy but appeared to improve conversion. 
We increased the numbers of channels of the initial convolution and the subsequent layer-groups 
from 64,64,128,256,512 to 128,256,512,1024,2048. Note that the dimension of the feature space matches that 
of resnet50 and the output dimension of the output of the initial convolution almost matches the input dimension. 
The network r18U achieved a validation accuracy of over 76.5\%, compared to resnet50's 75.1\%, in our pytorch setup. 
Network r18U is less efficient than resnet50, it has significantly more parameters.
Larger depth increases the potential number of pieces in the piecewise linear function for a similar number of 
neurons. 

Our algebraic analysis and experimental results as well as theory (two layer theorem) suggest that roughly the same 
final accuracy can be achieved by many network families, the distinguishing factor is the required model size. 
Imposing structure via convolutions, higher depth, residual blocks and their substructures increases 
the granularity of the piecewise linear function for a network with a given number of neurons.

\section{Polyharmonic Spline Neural Architecture Search} \label{sec:NASspline}
Given the insights from the prior Section, we developed a method to efficiently investigate a high dimensional parameter search space. We analyzed the search space of  a single network family like {\em ResNet} or {\em ResNeXt}. Widths and depths of blocks already provide 8 dimensions within any deep network family search space. On top of those, parameters internal to the blocks, as for example cardinality in {\em ResNeXt},
 would increase the number of dimensions even further, but based on the algebraic interpretation were likely to have less impact on final accuracy. 

Three key issues make NAS challenging for larger classification problems such as ImageNet1k or ImageNet22k: search (and evaluation) time, memory consumption, and probability of getting stuck in local minima.

First, in order to obtain a measure for the final accuracy of a model, it has to be trained nearly to saturation, that is, over a sufficiently large number of epochs to ensure that all the variants tested reach close to their full potential. In our experiments, early stopping proved to be misleading, since smaller networks tend to initially converge faster and the crossover point was near the end of a complete 
  learning rate schedule. Training ImageNet1k for 90 epochs and ImageNet22k for 60 epochs (values experimentally shown to provide meaningful accuracies) requires large amounts of computation. Hence, as in most NAS approaches, {\em minimizing the number of evaluations is critical}.
  
Second, the algebraic observations suggested that the better performing networks are large, such that GPU {\em memory limitations become a factor}. ``Supernet'' approaches such as FBNet\cite{wu2018fbnet} would not fit even two variants of the larger and more performing networks into a 16 GB GPU. In fact, the most accurate networks on ImageNet22K tend to use most of the memory of even a 32 GB GPU. 
  
Third, the topologies of the hypersurfaces defined by the parameter dimensions and the achieved final accuracy as 
  the evaluation axis may have local minima, i.e. {\em gradient descend based methods are suboptimal to find minima in the accuracy-hypersurface}.

Importantly, the algebraic investigation suggests that small changes to parameters such as the number of filters or 
number of layers in a ``convolution group'' or cardinalities of elements of basic blocks within a ``network family'' like ResNeXt or BLResNeXt will result in small changes in the final accuracy, since they cause only small changes in
the degrees of freedom in the number of pieces in the piecewise linear function and the equations that govern 
the pieces. Our experimental results support this hypothesis. Hence, it is appropriate to assume that there exists a set of 
continuous functions in $f : R^{d} \rightarrow R$ that pass through a set of reasonably spaced support points over $d$ discrete 
parameters and the resulting validation accuracy after training the network (close) to saturation. 

\subsection{Polyharmonic Splines} \label{ssec:spline}

Polyharmonic splines are an interesting option to define a function that passes through such a set of support 
points. Given a set of $N$ {\em support points} $(\vec{x}_j,f(\vec{x}_j))$ for a function $f:R^d \rightarrow R$,
  a polyharmonic spline interpolation has the form 
  \[
  s(\vec{x}) = \sum_{j=1}^N c_j \Phi(|| \vec{x} - \vec{x}_j ||) + p(\vec{x}) 
  \]
  where $||\vec{x}||$ is the Euclidian norm in $R^d$ and $p(\vec{x})$ is a real valued polynomial 
  in $d$ variables \cite{book_iske}. $\Phi$ is a radial basis function.  

The polyharmonic radial basis functions are solutions to a polyharmonic equation, a partial differential equation
of the form $\Delta^m f$ = 0. Polyharmonic splines minimize the ``curvature'' of the hyperplane that passes 
through all the support points, hence they minimize oscillations while providing a smooth, continuous surface.
This interpolation expression is differentiable. 
Thus, if we assume there exists a continuous and differentiable function $f:{\cal{R}}^{d} \rightarrow \cal{R}$ 
that correctly models the behavior of the system that generates the support points, then there exists a polyharmonic spline 
such that the integral of the difference between spline and $f$ in an interpolation {\em d-box} vanishes as the 
number of support points increases. 

We chose a radial basis function of $\Phi(r) = r^3$. Solving the equation system to determine the coefficients for the polyharmonic spline 
proved to be sensitive to numerical instability. Hence we chose a pivoting Householder QR decomposition (the EIGEN implementation), trading 
performance for numerical stability. 

In order for the linear equation system that determines the spline coefficients to have a solution, the matrix formed by 
the support points has to have full rank. Since interpolation accuracy is in general higher than extrapolation accuracy, 
a minimal set of support points it needed that spans a {\em d-box} in the d-dimensional space in which the spline interpolates 
the approximated function which leads to a support point matrix of full rank. We chose the following set for $d$ parameters: 
\begin{itemize}
  \item $\{max_d | d=1..N \}$, the vector of the largest usable or legal parameter values  in the {\em d-box} for all dimensions.
  \item $\{min_d | d=1..N \}$, the smallest usable or legal parameter values. 
  \item $~ \{ (max_d+min_d)/2 | d=1..N \}$, a point near the center point of the {\em d-box}. 
  \item $\{ max_d | d \neq k, d \in \{1..N\}, min_d | d==k\}$ for all $k=1..N$.
  \item $\{ min_d | d \neq k, d \in \{1..N\}, max_k | d==k\}$ for all $k=1..N$.
\end{itemize}
For two dimensions, these are the corners of a square and its center. For three dimensions, these are the corners of 
a cube and its center. 
With these support points, a total of $2d+3$ points is sufficient to span an initial cubic spline that interpolates within a {\em d-box}. 
Additional support points can be added to improve the quality of the interpolation. To maintain numerical stability, new support points 
have to maintain a certain distance from previous support points. We eliminated splines due to numerical instability 
by checking $ Ax-B = 0$ on the solution with an error margin for floating point computation errors.

\subsection{Minima Search}

  Since gradient descent is susceptable to local minima, we used a hierarchical Monte Carlo approach to 
search for a minimum in the interpolation d-box, with a sequence of low discrepancy, specifically a Halton 
sequence. For a given d-box, reducing the average distance between sample points by a factor of two 
requires an exponential increase in additional sample points, i.e. progress in terms of finding better minima 
slows daramatically as more sample points are added. 

As the average distance between sampling points decreases relative 
to the distance between support points, the curvature minimizing property of the polyharmonic spline reduces the risk 
of missing minima. Hence a hierarchical approach, iteratively shrinking the d-box around the minimum found so far, 
significantly reduces the time to determine a good approximation for the minimum within the d-box. The search was stopped 
if no further improvement within floating point accuracy was achieved within a certain compute budget. 

Note that in our architecture search a potentially bad estimate for a minimum does not limit progress of the search, as described in Algorithm \ref{alg:spline}. 
It merely leads to a potentially suboptimal measurement point and hence to potentially requiring more measurement points 
to reach convergence between prediction and measurement. 

\begin{algorithm}[H]
\SetAlgoLined
\KwIn{\\(a) Initial set of $2d+3$ support points $\vec{x}=\{\bm{x}_1,\bm{x}_2,...\}$, where $d$ is the number of dimensions of each point $\bm{x}_i$\;
(b) Measured function values $\vec{y}_m = f_m(\vec{x})$ for initial points\;
(c) Minimum Difference $\epsilon$ between prediction ($f_s$) and measure ($f_m$)\;}
\Begin{Solve equation system for spline $f_s(\vec{x})$ over initial support points\;
  Compute $\bm{x}_{max} = \underset{\bm{x}} {\arg\max} f_s(\vec{x})$ via nested Monte Carlo Sampling\;
  
  Compute measurement $f_m(\bm{x}_{max})$ by training network to saturation\;
  $\bm{x}_{top} = \bm{x}_{max}$\;
  
  \While{$|f_m(\bm{x}_{max}) - f_s(\bm{x}_{max})| > \epsilon$}{
  Add $\bm{x}_{max}$ to set of support points $\vec{x}$\;
  Solve equation system for spline $f_s(\vec{x})$ over all support points\;
  Compute $\bm{x}_{max} = \underset{\bm{x}} {\arg\max} f_s(\vec{x})$ via nested MCS\;
  Compute measurement $f_m(\bm{x}_{max})$ by training network\;
  \If{$f_m(\bm{x}_{max}) > f_m(\bm{x}_{top})$}{
    $\bm{x}_{top} = \bm{x}_{max}$\;
  }
 }
 }
 \KwResult{Optimal Parameters Configuration $\bm{x}_{top}$}
 \caption{Polyharmonic Splines NAS}
 \label{alg:spline}
\end{algorithm}

\section{Experiments} \label{sec:experiments}
In our experiments we investigated a couple of micro (within basic block structure) and multiple macro (overall network dimensions) parameters for two network families, namely ResNet \cite{resnet} and Big-Little ResNeXt \cite{blresnext}. Search was performed directly on the target dataset ImageNet22K. Training experiments were conducted on the the Summit Supercomputer at Oakridge National Lab, using 34 nodes each equipped with 6 Nvidia Volta V100 GPUs with 16GB GPU cache each. for a total of 204 GPUs. All GPUs in a node have NVLink connection, and the nodes are  connected by Mellanox EDR 100G Infiniband and have access to shared GPFS storage. Software used included PowerAI Vision 1.6, NCCL and Pytorch, using its distributed data parallel package. Batch size was set at 32 per GPU, for a total batch size of 6,528. The initial learning rate was set at 0.1 and the followed polynomial decay, optimizer SGD with momentum 0.9 and weight decay 0.0001.
 
\subsection{Experimental Dataset: ImageNet22k}
\begin{figure}[t]
   \centering
    \begin{subfigure}{0.48\linewidth}
        \includegraphics[width=\linewidth]{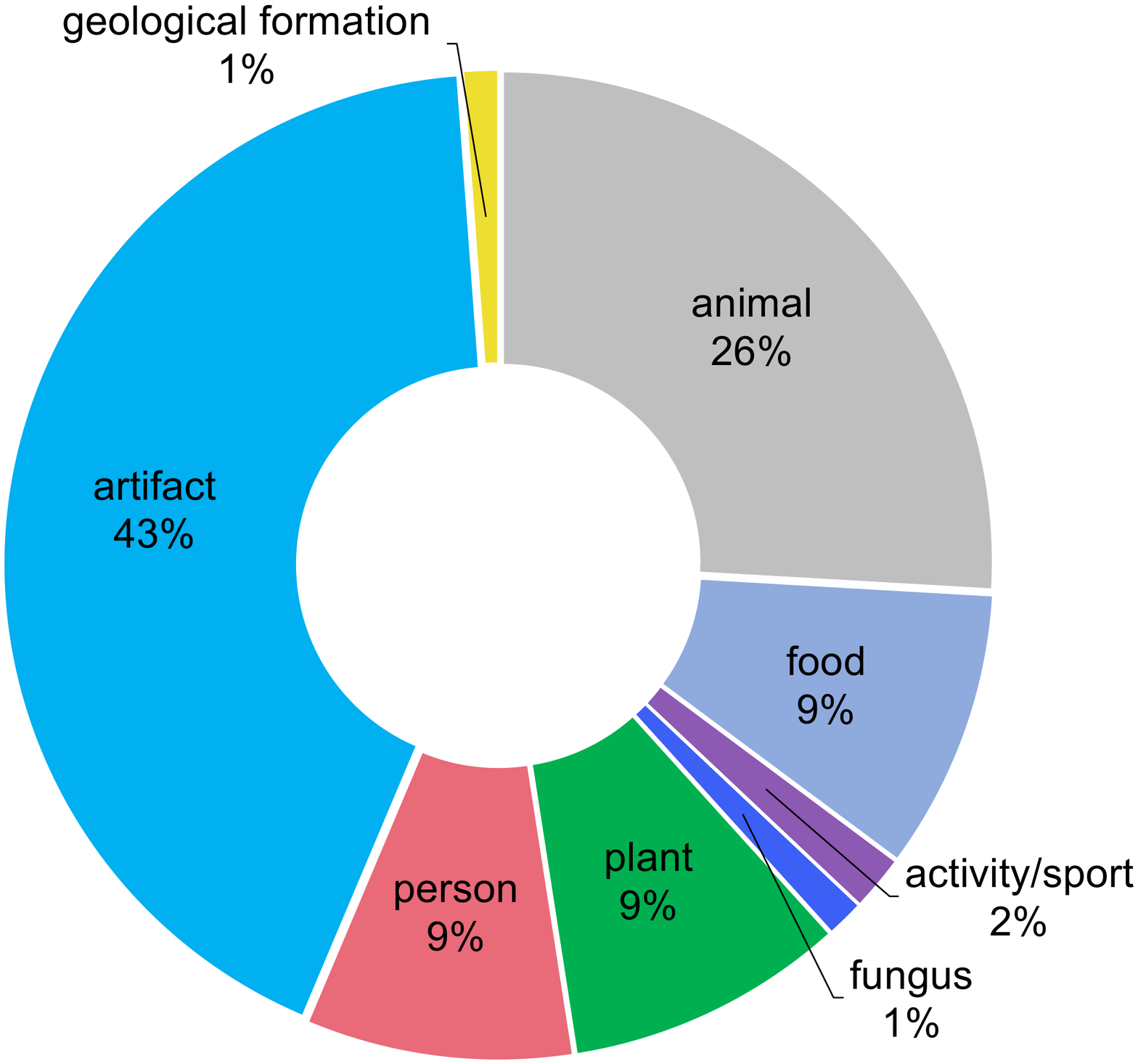}
        \label{fig:im22k_top_dist}
    \end{subfigure}%
    \begin{subfigure}{0.52\linewidth}
        \includegraphics[width=\linewidth]{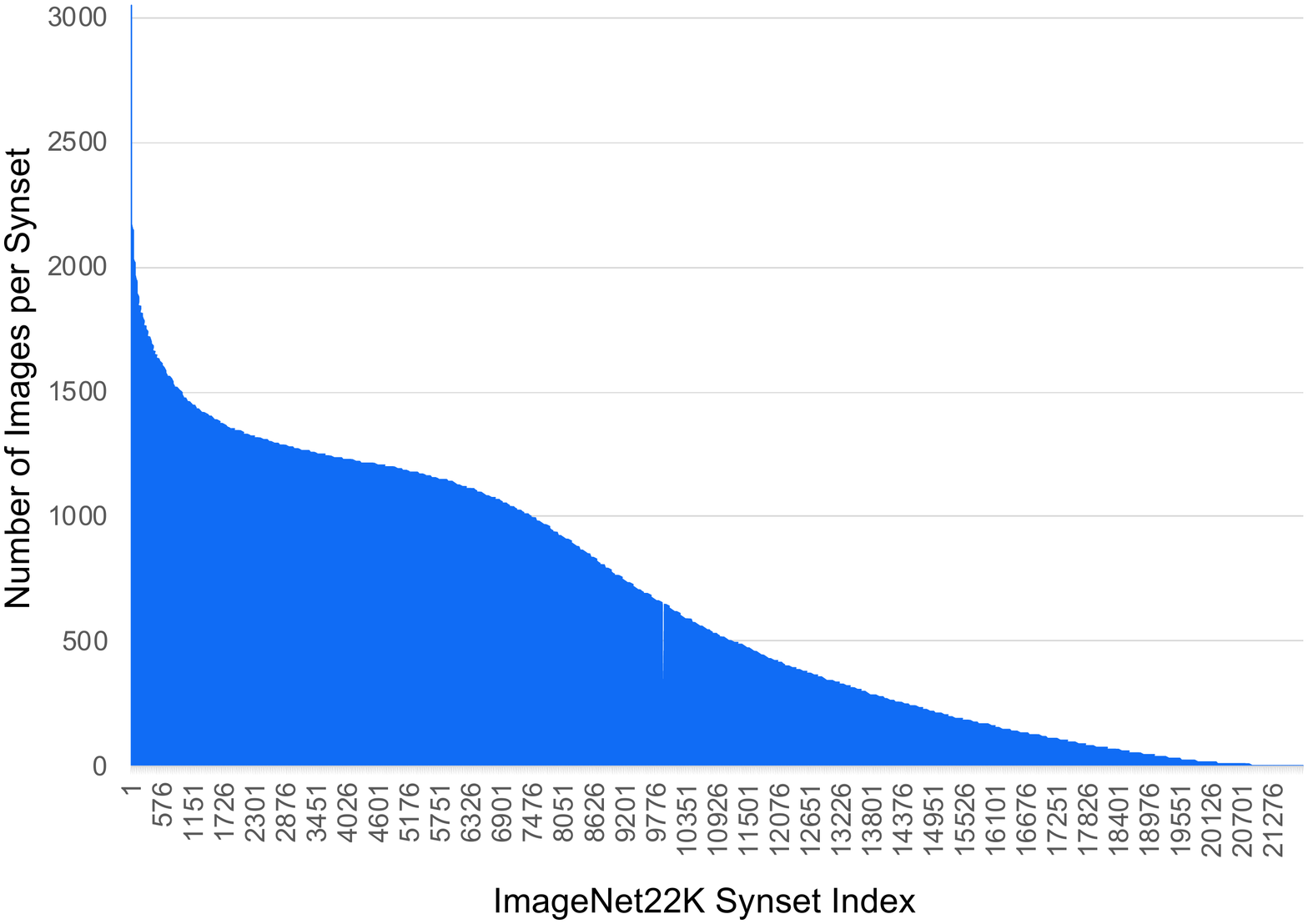}
        \label{fig:im22k_all_dist}
    \end{subfigure}%
   \caption{ImageNet22K taxonomy of higher level labels (left) and distribution of images per label (right).}
   \label{fig:22kimgs}
\end{figure}

ImageNet22k contains 14 million images representing 21,841 categories organized in a  hierarchy derived from WordNet and including top level concepts such as sport, garment, fungus, plant, animal, furniture, food, person, nature, music, fruit, fabric, tool etc. Figure \ref{fig:22kimgs} shows the top level catefories of the ImageNet22k hierarchy dataset and their relative sizes in terms of number of images. The imbalance across top level semantic categories is quite evident. For example animal is represented three times as much as person, and artifacts dominate the distribution with very little representation for activities. The skew is also significant in terms of number of examples per category. On average there are images per class, ranging from a minimum of to a maximum of. The scale and imbalance of ImageNet22K make it particularly challenging even for human designed architectures, with a limited set of published results~\cite{adamOSDI2014,cho2017powerai,intel_blog_2017,zhang2015poseidon}, as opposed to the smaller and balanced ImageNet1K version. Recently, some works have used ImageNet22K as pre-training for Imagene1K evaluation \cite{anonymous2021an}. To the best of our knowledge, this is the first work to perform NAS directly at the scale of ImageNet22K, not by transfer from smaller proxy sets.  

Following standard practise~\cite{bhattacharjee2017ibmjr,adamOSDI2014,deng2010eccv} we randomly partitioned the ImageNet22K dataset into 50\% training and 50\% validation, consisting of approximately 7 million images each. We split the data into two sets such that number of images per label are approximately equal in both sets. In the cases where a label had odd number of images, we put the extra image in the validation set. 

\subsection{Results} 
We applied our polyharmonic spline NAS method to the the ResNet18 and BLResNext50 architectures search spaces. For each point in the spline, training and evalution was conducted on half of the ImageNet22k dataset. Once the optimal configuration was determined by our search, that network was trained and evaluated on the full ImageNet22k dataset.

\begin{table}[ht]
\begin{center}
\captionof{table}{Simplified ResNet18 with 15 points over 5 dimensions of search including 13 points for initial spline and 2 incremental points}
\label{tbl.tabres18.1}
\resizebox{\textwidth}{!}{
\begin{tabular}{c|c|c|c|c|c|c|c}
\toprule
\multirow{2}{*}{\textbf{Point Type}} & \multicolumn{5}{c|}{\textbf{Dimensions}} &  \multicolumn{2}{c}{\textbf{Top-1 Accuracy \%}}  \\
\cline{2-8}
 &\textbf{conv \textit{c1}} & \textbf{group \textit{g1}} & \textbf{group \textit{g2}} & \textbf{group \textit{g3}} & \textbf{group \textit{g4}} & \textbf{Measured } & \textbf{Predicted} \\ 
 \hline 
 \multirow{13}{*}{Initial} & 150 &  300 &  600 &  1200 &  2400 &  38.03 & - \\
 & 32 &  32  &  32  &  32   &   32  &  10.33  &  -\\
 & 150 &  32  &  32  &  32   &   32  &  10.54  & - \\
 & 32 & 300  &  32  &  32   &   32  &  15.46  &  - \\ 
 & 32 &  32  & 600  &  32   &   32  &  19.45  &  - \\
 & 32 &  32  &  32  & 1200  &   32  &  23.15  &  -  \\
 & 32 &  32  &  32  &  32   &  2400 &  31.25  &  -  \\
 & 75 & 150  & 300  & 600   &  1200 &  35.46  & -   \\
 & 32 & 300  & 600  & 1200  &  2400 &  38.03  &  -  \\ 
 & 150 &  32  & 600  & 1200  &  2400 &  37.15  & -   \\
 & 150 & 300  &  32  & 1200  &  2400 &  36.76  &  -  \\
 & 150 & 300  & 600  &  32   &  2400 &  34.40  & -   \\
 & 150 & 300  & 600  & 1200  &   32  &  29.04  & -   \\ \hline
 \multirow{2}{*}{Incremental} & 50  & 116  & 330  & 1200  & 2400  &  37.31  & 41.02\\
 & 80  & 208  & 475  & 736   &2400   &      -      & 38.92\\ 
 \bottomrule
\end{tabular}%
}
\end{center}
\end{table}

\subsubsection{ResNet18 Search Space.}
For the ResNet18 architecture we removed maxpool layer and increased the stride for the first convolution to $2$. 
We performed search over $d=5$ dimensions $c1,g1,g2,g3,g4$: the number of filters in the first convolution ($c1$) and in the four groups of layers ($g1,g2,g3,g4$).
The ranges for each dimension are as follows: $c1\in[32,150]$, $g1\in[32,300]$, $g2\in[32,600]$, $g3\in[32,1200]$ and $g4\in[32,2400]$.
Therefore the possible combinations spanning the entire search space are $\sim$3 trillion ($120\times135\times570\times1170\times2368$). Using our spline NAS approach allows to start from \textit{only} an initial set of 13 support points ($2d + 3$, as explained in Section~\ref{ssec:spline}) and then iterate from there with few additional points. This results in a tremendous computational gain for our NAS method which allows to perform search directly on large scale datasets, as opposed to traditional NAS approaches needing to resort to small proxy sets.

Table~\ref{tbl.tabres18.1} shows the coordinates for the initial 13 support points to span the first spline and two incremental points. An incremental point 
is a measurement on the optimum estimated by the prior set of points. The first prediction, based on the minimal point set, predicted a reasonable 
 point, but the estimate of $41\%$ is clearly very optimistic. Adding a measurement 
 at this point produces a new point with a prediction close to the best support point.
 
\begin{figure}[t]
   \centering
    \begin{subfigure}{0.49\linewidth}
        \includegraphics[width=\linewidth]{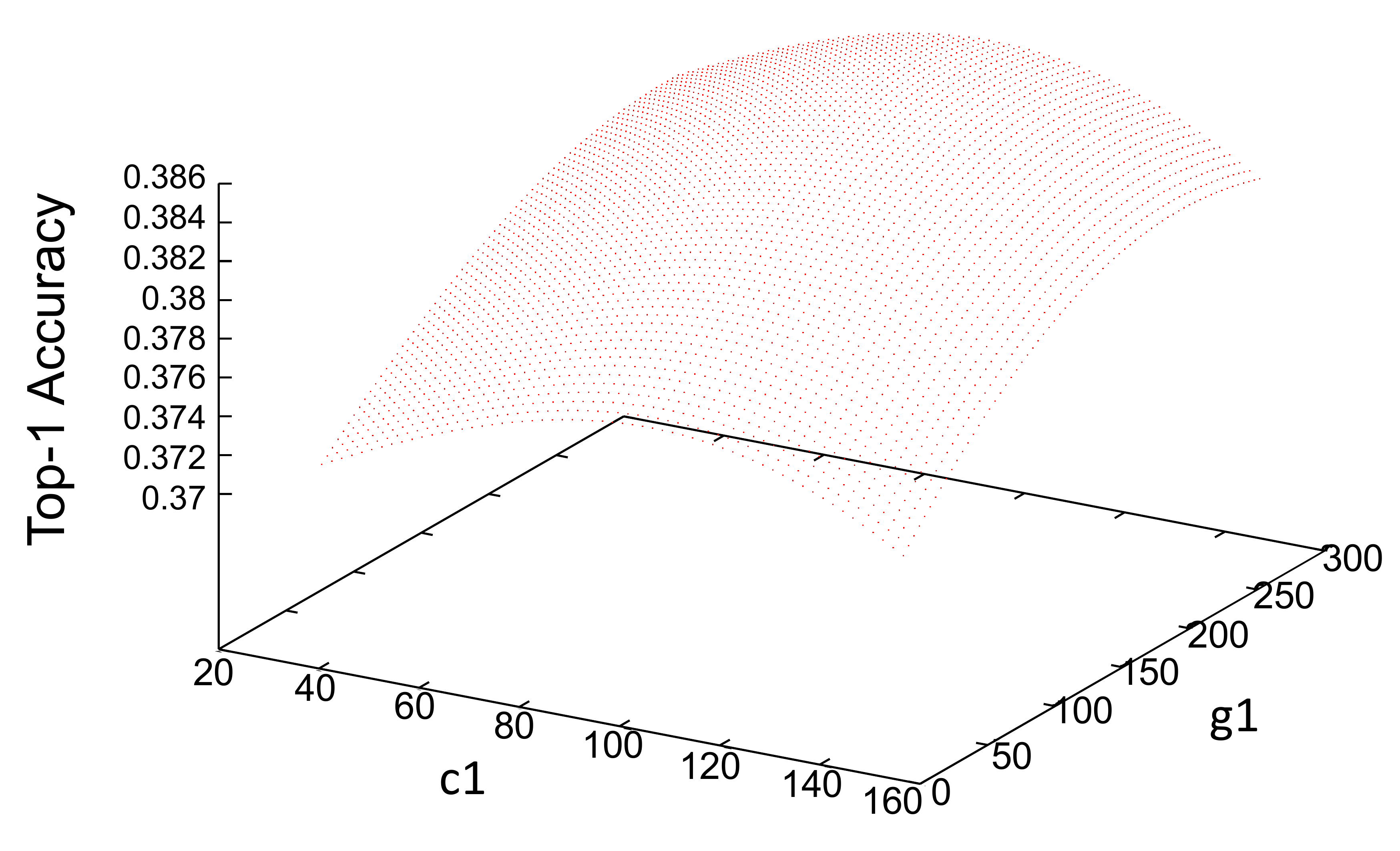}
        \caption{Projection $[c1,g1,600,1200,2400]$}
        \label{fig-proj_r18_a_b}
    \end{subfigure}
    \begin{subfigure}{0.49\linewidth}
        \includegraphics[width=\linewidth]{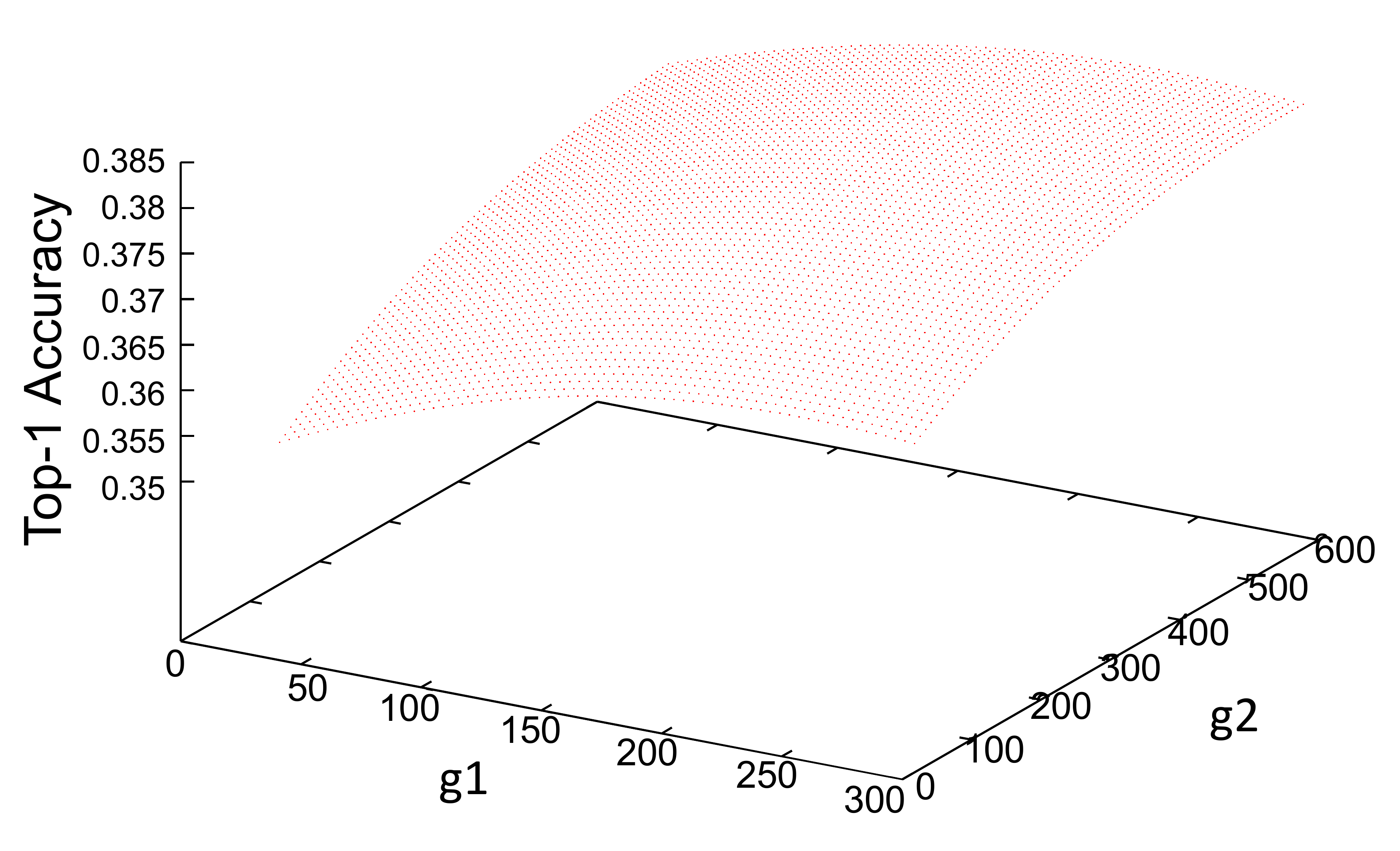}
         \caption{Projection $[150,g1,g2,1200,2400]$}
        \label{fig-proj_r18_b_c}
    \end{subfigure}
    \begin{subfigure}{0.49\linewidth}
        \includegraphics[width=\linewidth]{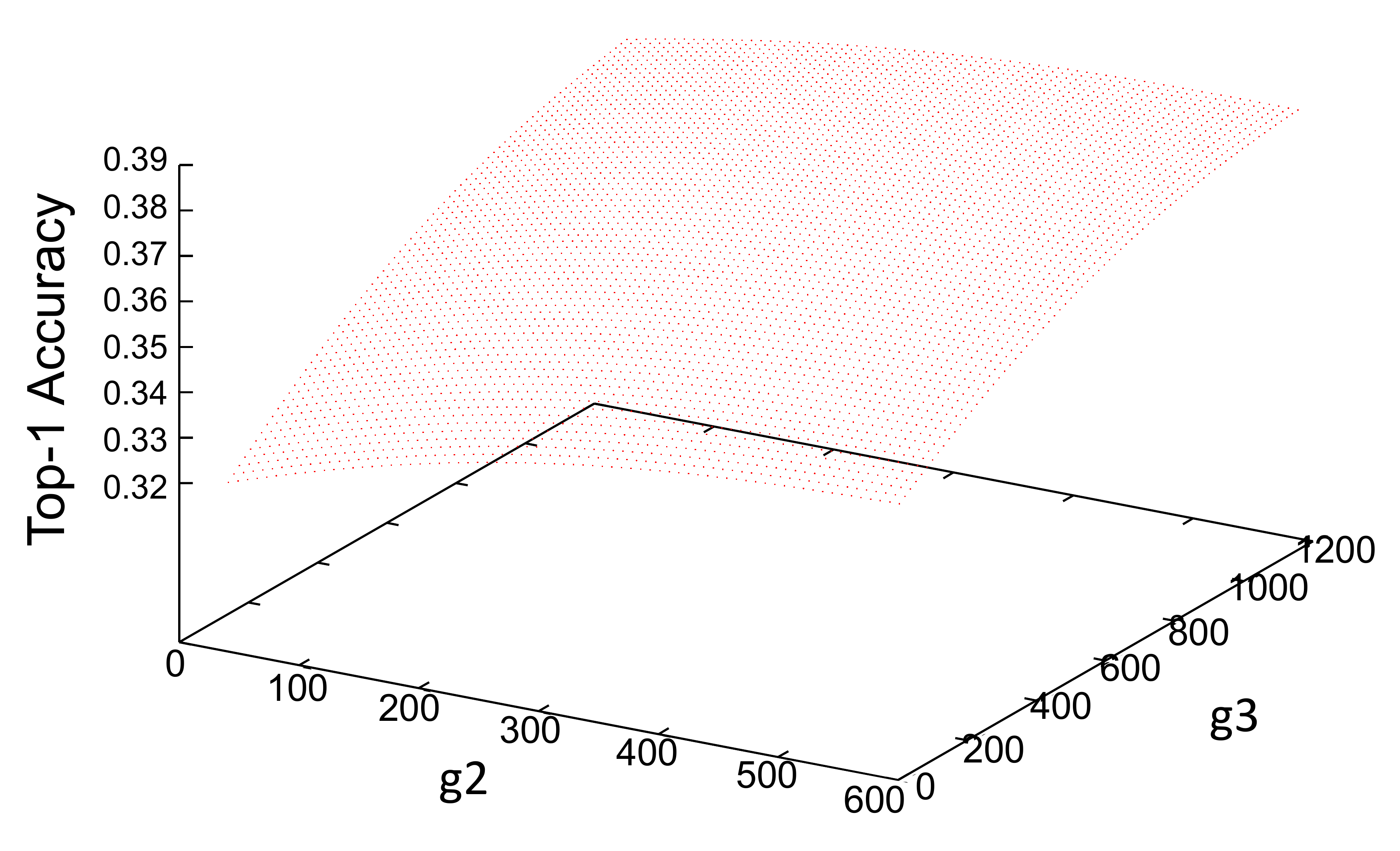}
        \caption{Projection $[150,300,g2,g3,2400]$}
        \label{fig-proj_r18_c_d}
    \end{subfigure}
    \begin{subfigure}{0.49\linewidth}
        \includegraphics[width=\linewidth]{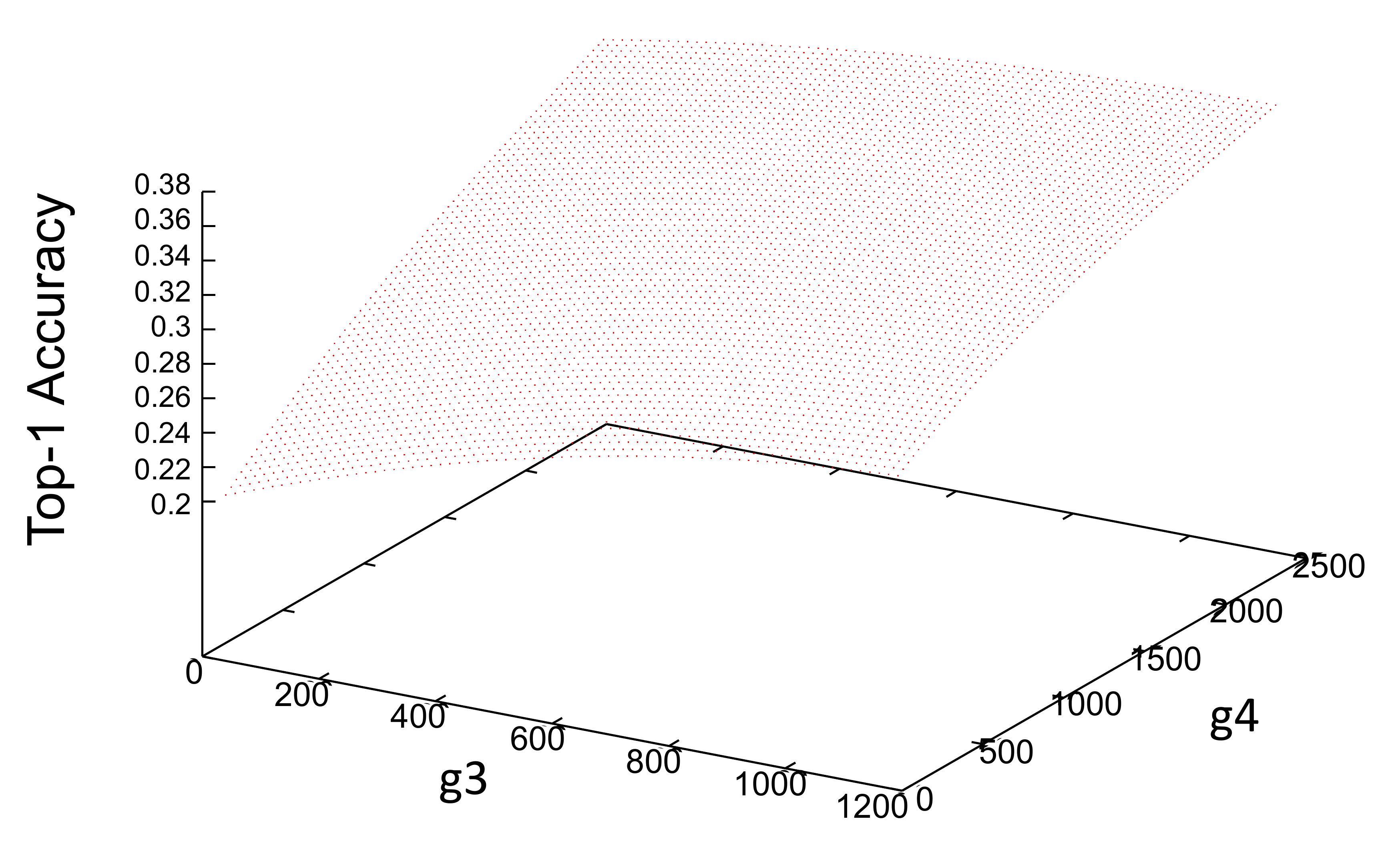}
         \caption{Projection $[150,300,600,g3,g4]$}
        \label{fig-proj_r18_d_e}
    \end{subfigure}
   \caption{Top-1 accuracy on ImageNet22k (half) for the ResNet18 architecture by projecting the search space to two dimensions. For each projection, three parameters are fixed and performance is inspected over the search space spanning the range of the remaining pair of parameters: (a) varying $c1$ and $g1$, (b) varying $g1$ and $g2$, (c) varying $g2$ and $g3$, (d) varying $g3$ and $g4$.}%
    \label{fig:projections}

\end{figure}

  Figure~\ref{fig:projections} shows projections of the polyharmonic spline derived from the $14$ measured points as show in Table~\ref{tbl.tabres18.1}. 
The interpolation suggests that the parameter $d5$ is the dominant limiting factor, since it shows the steepest slope at the edge of the ``box''. This matches the algebraic interpretation, $\sim$22k classes could benefit from a higher dimensional feature space. 
The earlier layers/group show maxima within the box for maximum values for the 
later layers, indicating that once the degrees of freedom of a later part of the 
network are saturated, adding more capacity to earlier layers becomes counterproductive.





Hence, we measured configuration of [$300,600,1200,2400,5400$], roughly projecting out the ratios 
of the optimum from the spline with some adjustments to fit into available GPU memory. 
This achieved, all other hyperparameters remaining equal, a top-1 accuracy of 
$39.76\%$. Since this network was significantly larger and hence may benefit from 
a different learning rate schedule, we performed 2 epochs of fine tuning, which 
increased the accuracy to $40.37\%$. 

Being limited by GPU memory and compute resource, we performed one more experiment 
to increase the number of pieces in the piecewise linear function without increasing 
the model size significantly by replacing the $7\times7$ convolution with stride 2 at 
the beginning with a basicblock. A basicblock consists of two $3\times3$ convolutions, one of  which has stride 2, which has a similar aperture and the same reduction of the activations. Indeed, this yielded the expected improvement, $40.68\%$ top-1 accuracy with fine-tuning. 

\subsubsection{BLResNext50 Search Space} 

Big-Little Net\cite{blresnext} is a mechanism that splits each block within a deep network into multiple paths through which different resolutions of an input are passed. In our search space we considered two paths. The first, called big branch, through which a downsized (by half) version of the input is passed, containing $C$ kernels and $L$ layers. The second, called little branch, processing the input in its original resolution, but containing $C/\alpha$ kernels and $L/\beta$ layers.
  The big-little version of ResNext \cite{blresnext} with a depth of 50 is deeper and offers more alternate paths through groups than the basic ResNet18, and hence theoretically allows for more pieces in the piecewise linear function relative to the number of network parameters. Thus, this family of  networks promises to achieve higher accuracy within a given GPU memory capacity, which  was clearly the limiting factor for the ResNet18 case. 

we defined our search space as spanning only three variables, hence reducing the number of needed measurements for optimization. We chose as parameters the $\alpha\in[2,8]$ and $\beta\in[2,8]$ parameters of the big-little structure and a multiplier $\phi\in[1.0,2.0]$ to the group width, that gets uniformly applied across the network. The original group widths for BLResNext50 were $64,128,256,512$. With a bottleneck expansion factor of $4$, this results in a 2048-dimensional feature space. The choice $\phi = 2$ for example results in group width $128,256,512,1024$ and a 4096-dimensional feature space. The total combinations of the search space are at least 539 ($7\times7\times11$, considering $\phi$ only at discrete increments of 0.1), but the spline optimization needs only $2\times3+3=9$ supporting measurements followed by a couple of additional ones.
 

 \begin{table}[t]
 \caption{BLResNext50, 3 dimensions ($\alpha$, $\beta$, and $\phi$), 9 points for initial spline, 1 incremental}
\label{tbl.tab-blr50}
\begin{center}
 \begin{tabular}{c|c|c|c|c|c|c|c|c|c|c}
\toprule
\textbf{Point type} & \multicolumn{9}{c|}{Initial} & Incremental\\
\hline
$\bm{\alpha}$   & 8       & 2       & 2       & 8       & 4       & 2       & 8       & 2       & 8   & 2    \\ 
$\bm{\beta}$    & 8       & 2       & 8       & 2       & 4       & 2       & 2       & 8       & 8    & 8   \\ 
$\bm{\phi}$  & 1       & 1       & 1       & 1       & 1.5     & 2       & 2       & 2       & 2  & 3     \\ \hline 
\textbf{Top-1 Accuracy \%}  & 38.17 & 38.83 & 38.75 & 38.18 & 39.90 & 40.96 & 40.53 & 40.99 & 40.48 & 41.64 \\ 
\bottomrule
\end{tabular}%
\end{center}
\end{table}


\begin{figure}[ht]
\centering
  \centering
  \includegraphics[width=0.6\linewidth]{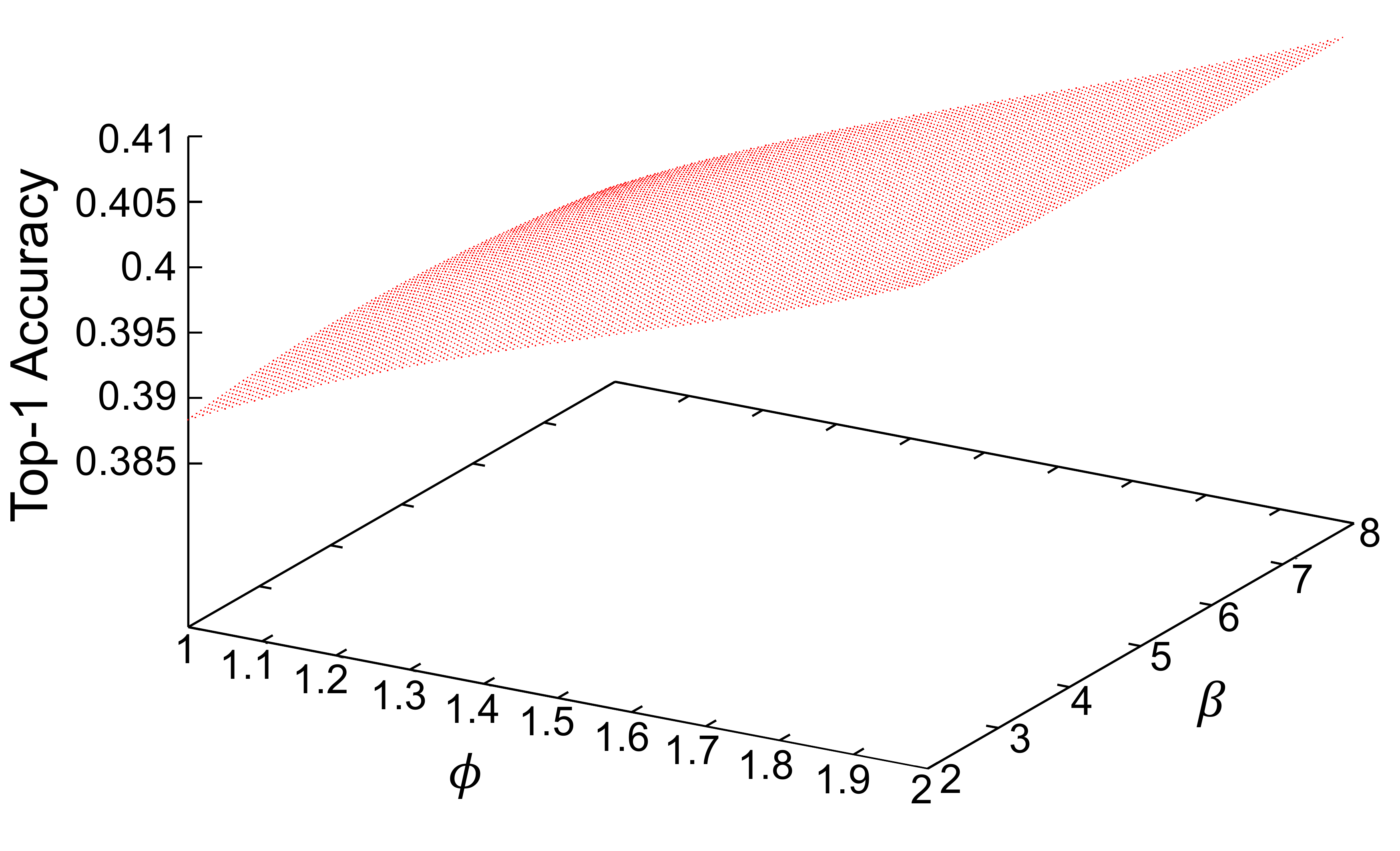}
  \captionof{figure}{Top-1 accuracy on ImageNet22k (half) for BLResNext50. Projection $\alpha=2,\beta,\phi$, accuracy over parameters $\beta$ and $\phi$.}
  \label{fig-blr50}
\end{figure}

\begin{figure}[ht]
  \centering
  \includegraphics[width=\linewidth]{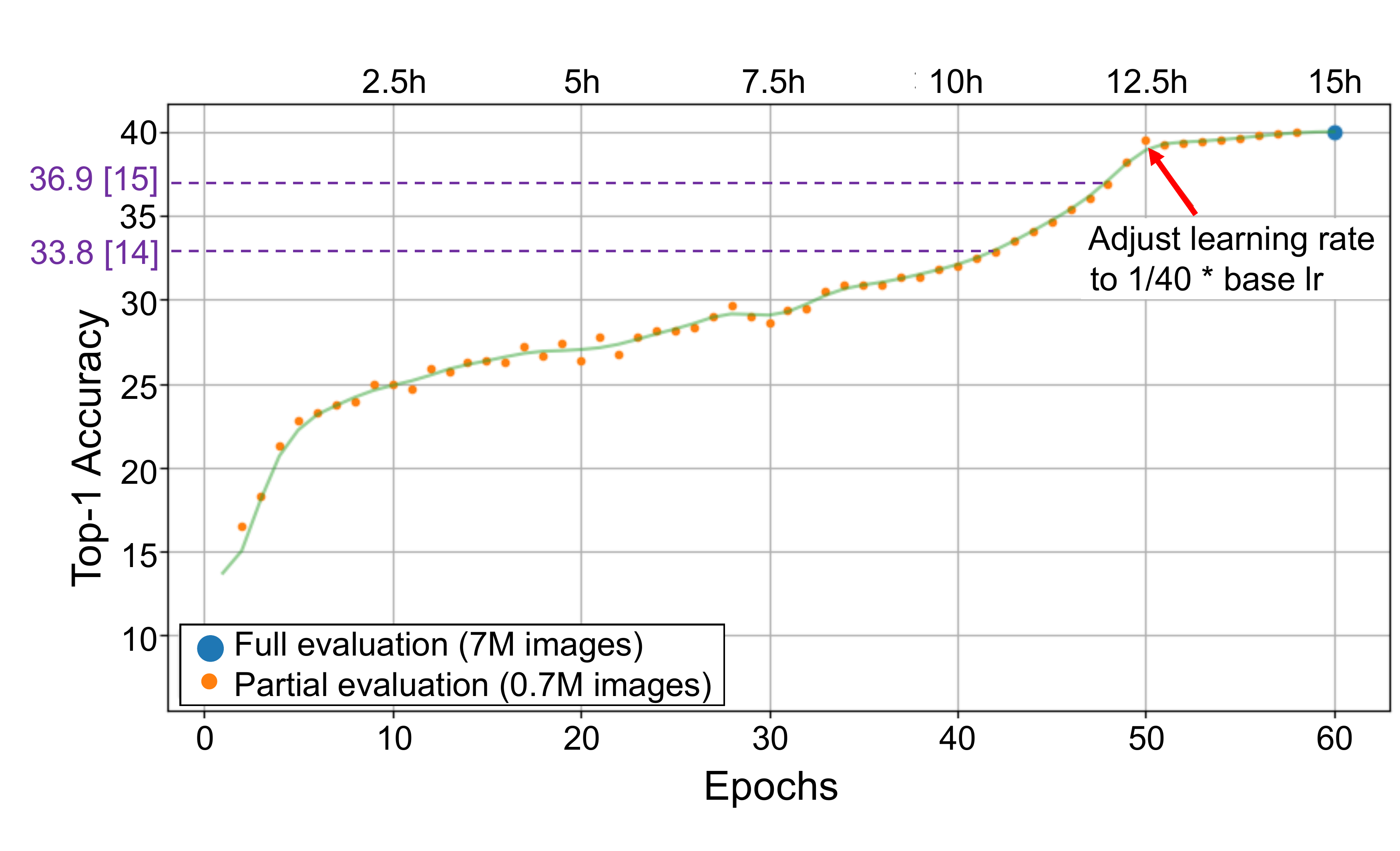}
  \captionof{figure}{BLResNext50 18x8d testing accuracy on the full ImageNet22K ($34$ nodes, $204$ GPUs, $6,528$ batch size on Summit Supercomputer.}
  \label{fig.22ksummit1}
\end{figure}




Table \ref{tbl.tab-blr50} shows that $\alpha$ and $\beta$ have only a small
influence on accuracy compared to $\phi$. Figure \ref{fig-blr50} shows the projection
$acc(\alpha=2,\beta,\phi)$. The dependencies within the ``box'' are nearly linear and 
the minimum is located in the $\alpha=2,\beta=8,\phi=2$ corner, clearly 
indicating that an increase in width has the best probability to increase accuracy 
significantly. Hence, we measured $\alpha=2,\beta=8, \phi=3$, which indeed yielded to top-1 accuracy of $41.64$. This was interesting also because the shape of the spline suggested an increment in the variable $\phi$ beyond the initially designed range of the search space.

The total amount of single Nvidia Volta GPU hours needed for the NAS Spline search was approximately $30,000$.  This is the accumulation of conducting evaluation (training and validation) half of the ImageNet22K dataset for all the configurations corresponding to the initial $2*3+3=9$ points and additional 3 data points. Each individual point measurement took about $2,500$ GPU hours. For reference, the original reinforcement learning based NAS~\cite{NAS_RL_ICLR2017} method would require a minimum of $539\times2,500=13.47M$ GPU hours. 

The final recommended BL-Net architecture was trained and evaluated on the full ImageNet22K dataset using the Summit supercomputer over $34$ nodes with $204$ Volta GPUs with a global batch size of $6,528$ using Pytorch distributed data parallel. Figure \ref{fig.22ksummit1} shows how the top-1 accuracy climbed as the learning progressed. On the way to the $40.3\%$ accuracy it crossed two previously published results as shown. 
In table \ref{tab:22ksummit2} the comparison of our result against previous results as well as a baseline SME designed architecture based on the BL-ResNext are summarized.  The SME designed architecture was a BL-ResNext 101 based model in comparison to the BL-ResNext 50 based Spline recommended model. As can be seen, the Spline recommended architecture resulted in a jump of $0.33\%$ increase in top-1 accuracy. This is the first published result which has crossed $40\%$ in overall top-1 accuracy on ImageNet22K. 




\begin{table}[t]
\caption{Comparison of ImageNet22K results. The last two row denotes the Spline NAS recommended architecture. FLOPs are estimated with a network forward pass using input  image resolution of 256x256}
\label{tab:22ksummit2}
\centering
\resizebox{\textwidth}{!}{
\begin{tabular}{c|c|c|c|c|c|c}
 \hline
  \multirow{2}{*}{\textbf{Model}} & \textbf{Batch} & \multirow{2}{*}{\textbf{GPUs}}  &\textbf{ Top-1(Top-5)} & \textbf{FLOPs} & \textbf{Training}  & \textbf{Number of}    \\ 
  & \textbf{Size} & & \textbf{Accuracy \%} & \textbf{(G)} & \textbf{Time(Hours)} & \textbf{Epochs} \\ \hline \hline
   ResNet-101 \cite{cho2017powerai} & 5,120 & 256 & 33.8 (-) & - & 7 & - \\ 
   WRN-50-4-2 \cite{intel_blog_2017} & 6,400 & 200 & 36.9 (65.1) & - & - & 24 \\
   BL-ResNext101 32x8d  & 6,528 & 204 & 39.7 (68.3) & 11.25 & 16 & 60 \\
   \textbf{BL-ResNext50 18x8d (ours)} & \textbf{6,528} & \textbf{204} & \textbf{40.03 (69.04)} & \textbf{17.88} & \textbf{15} & \textbf{60}  \\
   \hline
\end{tabular}
}
\end{table}



\subsection{BLResNext on ImageNet1K}

  In order to investigate the influence of 
  both group-width and group-depth in the Big-little-ResNext family, we picked 8 parameter dimensions
  and experimentally verified the convergence of the spline method using the smaller ImageNet1K dataset. The investigated
  parameters $\omega_1,\omega_2, \omega_3, \omega_4$ are the number of filters in the first layer in a layer group, i.e. the number of filters in the output of a group is increased by the expansion factor. Parameters $d_1,d_2,d_3,d_4$ are the depths, i.e. the number of calls to {\em make\_layer} 
  within a {\em basicblock} for the four block groups.

\begin{table}[h]
\caption{BLResNext, 8 dimensions, 19 points for initial spline. Check against known blresnext50 config ($64,128,256,512,3,4,6,3$) 
  followed by iterative points, measured top-1 accuracy vs prediction.}
\begin{center}
\begin{tabular}{c|c|c|c|c|c|c|c|c|c|c}
\toprule
\multirow{2}{*}{\textbf{Point Type}} & \multicolumn{8}{c|}{\textbf{Dimensions}} &  \multicolumn{2}{c}{\textbf{Top-1 Accuracy \%}}  \\
\cline{2-11}
 &  $\bm{\omega_1}$ & $\bm{\omega_2}$  &  $\bm{\omega_3}$ & $\bm{\omega_4}$ & $\bm{d_1}$ &  $\bm{d_2}$ &  $\bm{d_3}$ &  $\bm{d_4}$ &  \textbf{Measured} & \textbf{Predicted}  \\ \hline
  \multirow{19}{*}{Initial} & 32   &   32  &     32  &      32  &     2  &     2  &     2  &     2  &    52.95  &  -  \\
  & 128   &  256  &    512  &     768  &    10  &    10  &    18  &     5  &    78.82  &  -  \\
 & 80   &  160  &    320  &     480  &     5  &     5  &     5  &     3  &    77.61  &  -  \\
  & 128   &   32  &     32  &      32  &     2  &     2  &     2  &     2  &    60.60  &  -  \\
  & 32   &  256  &     32  &      32  &     2  &     2  &     2  &     2  &    65.53  &  -  \\
 & 32   &   32  &    512  &      32  &     2  &     2  &     2  &     2  &    70.19  & -   \\
 & 32   &   32  &     32  &     768  &     2  &     2  &     2  &     2  &    69.84  &  -  \\
 & 32   &   32  &     32  &      32  &    10  &     2  &     2  &     2  &    58.37  & -   \\
 & 32   &   32  &     32  &      32  &     2  &    10  &     2  &     2  &    58.83  &  -  \\
 & 32   &   32  &     32  &      32  &     2  &     2  &    18  &     2  &    59.17  &  -  \\
 & 32   &   32  &     32  &      32  &     2  &     2  &     2  &     5  &    55.16  &  -  \\
 & 32   &  256  &    512  &     768  &    10  &    10  &    18  &     5  &    77.68  &  -  \\
 & 128   &   32  &    512  &     768  &    10  &    10  &    18  &     5  &    78.18  &  -  \\
 & 128   &  256  &     32  &     768  &    10  &    10  &    18  &     5  &    78.26  &  -  \\
 & 128   &  256  &    512  &      32  &    10  &    10  &    18  &     5  &    77.78  &  -  \\
 & 128   &  256  &    512  &     768  &     2  &    10  &    18  &     5  &    78.55  &  -  \\
 & 128   &  256  &    512  &     768  &    10  &     2  &    18  &     5  &    78.40  &  -  \\
 & 128   &  256  &    512  &     768  &    10  &    10  &     2  &     5  &    78.44  &   - \\
 & 128   &  256  &    512  &     768  &    10  &    10  &    18  &     2  &    79.27  &  -  \\ \hline
  \multirow{3}{*}{Incremental}  & 71   &  256  &    512  &     768  &     2  &     2  &     2  &     2  &    76.42  & 85.09   \\
 & 91   &   72  &    512  &     768  &     6  &     7  &    10  &     2  &    77.76  & 80.88 \\
 & 128   &  256  &    441  &     768  &    10  &    10  &    18  &     2  &          -   & 79.29 \\  \hline
 BL-ResNext50 Default & 64   &  128  &    256  &     512  &     3  &     4  &     6  &     3  &    77.02  & 75.50 \\
  \bottomrule
\end{tabular}  
\end{center}
\label{tab:blr1k}
\end{table}

Table \ref{tab:blr1k} shows the 19 ($2\times8+3$) measured points that span an initial spline and two points with a 
comparison of the prediction against a measurement. After the initial set of observations, we run predictions of the spline on different points within the search space. As an example, we can look at the default BL-ResNext50 configuration (last row in the Table). Since it is located close to the center of the {\em d-box} and hence close to a support point, the relative error between the top-1 accuracy predicted by the spline model and the measured performance is moderate, approximately 1.5\% (77.02\% measured versus the predicted 75.50\%). 

The point $71,256,512,768,2,2,2,2$ is the minimum found within the {\em d-box}. It is located 
relatively far from the support points, hence its prediction is much less accurate. Iteratively adding 
minima as support points tends to quickly improve the accuracy of predictions and thus leads 
to good network parameters. If the prediction quality doesn't improve, adding more support points 
in the region of interest, e.g. spanning a smaller {\em d-box} inside the first one, can improve 
the splines predictive capabilities. 

Adding $71,256,512,768,2,2,2,2$ to the base spline (not including the default BL-ResNext50 point) delivers 
a new prediction, $91,72,512,768,6,7,10,2$ with predicted top-1 accuracy of $80.88\%$. Adding the measurement 
for this second point to the spline delivers a new prediction with accuracy $79.29\%$ which closely matches a 
measured point of the corner of the d-box. That suggests that this corner point is the optimimum within this d-box. 

The location of the minimum in a corner point suggests that better parameters may exist outside the interpolation 
d-box. Hence, we added a set of measurements to expand the d-box to a wider range of parameters for the widths 
of the convolution groups to start a new iteration. Table \ref{tab:blr1k_B} shows the additional points and 
the predictions and measurements. Not unsurprising, the point at the maximum edge of the new d-box shows already 
an improved final top-1 accuracy of 79.38\%. Additional iterative points close the gap between prediction and measurements.
We stopped the iteration as it became evident that still the best corner point would be very close to
a settled maximum. The iteration unveiled a smaller network with almost identical 
final accuracy of $79.36\%$ at $235,355,408,872,10,10,18,3$. 

\begin{table}[ht]
\caption{BLResNext, 8 dimensions, Additional points for widened spline. Includes blresnext50 config $64,128,256,512,3,4,6,3$ 
  and measured iterative points for the narrower d-box from table \ref{tab:blr1k}, measured top-1 accuracy vs prediction.}
\begin{center}
\begin{tabular}{c|c|c|c|c|c|c|c|c|c|c}
\toprule
\multirow{2}{*}{\textbf{Point Type}} & \multicolumn{8}{c|}{\textbf{Dimensions}} &  \multicolumn{2}{c}{\textbf{Top-1 Accuracy \%}}  \\
\cline{2-11}
 & $\bm{\omega_1}$ & $\bm{\omega_2}$  & $\bm{\omega_3}$ & $\bm{\omega_4}$ & $\bm{d_1}$ & $\bm{d_2}$ & $\bm{d_3}$ & $\bm{d_4}$ &  \textbf{Measured} & \textbf{Predicted}  \\ \hline
  \multirow{13}{*}{Initial} & 256 & 512 &  768 &  1024 &  10 &  10 &  18 &   5 &  79.38   &     -   \\
 & 256 &  32 &   32 &    32 &   2 &   2 &   2 &   2 &  64.91  &  -    \\
 & 32 & 512 &   32 &    32 &   2 &   2 &   2 &   2 &  68.82    &  -  \\
 & 32 &  32 &  768 &    32 &   2 &   2 &   2 &   2 &  70.92   &  - \\
 & 32 &  32 &   32 &  1024 &   2 &   2 &   2 &   2 &  70.51    &  -  \\
 & 32 & 512 &  768 &  1024 &  10 &  10 &  18 &   5 &  78.12   &  -  \\
 & 256 &  32 &  768 &  1024 &  10 &  10 &  18 &   5 &  78.71  &  -          \\
 & 256 & 512 &   32 &  1024 &  10 &  10 &  18 &   5 &  79.01   &  -           \\
 & 256 & 512 &  768 &    32 &  10 &  10 &  18 &   5 &  78.81   &  -   \\
 & 256 & 512 &  768 &  1024 &   2 &  10 &  18 &   5 &  79.10   & - \\
 & 256 & 512 &  768 &  1024 &  10 &   2 &  18 &   5 &  79.25    &   -\\
 & 256 & 512 &  768 &  1024 &  10 &  10 &   2 &   5 &  79.11  &  - \\
 & 256 & 512 &  768 &  1024 &  10 &  10 &  18 &   2 &  79.17   &    - \\ \hline 
 \multirow{9}{*}{Incremental} & 210 & 357 &  433 &   500 &   3 &   5 &  13 &   2 &  78.83   & 81.78	  \\
 & 235 & 355 &  408 &   872 &  10 &  10 &  18 &   3 &  79.37   & 80.13   \\
&  130 & 289 &  417 &   489 &   3 &   4 &   9 &   3 &  78.47   & 79.93    \\
 & 158 & 314 &  381 &   761 &   5 &  10 &  18 &   3 &  79.04   & 79.86   \\
 & 240 & 400 &  620 &   532 &  10 &   8 &  18 &   2 &  79.16   & 79.76   \\
 & 256 & 385 &  433 &  1023 &  10 &   3 &  18 &   5 &  79.31   & 79.6   \\
 & 188 & 365 &  445 &   875 &  10 &  10 &  18 &   2 &  79.18    & 79.51   \\
 & 256 & 293 &  363 &  1024 &  10 &  10 &  18 &   5 &  79.24   & 79.58   \\
 & 180 & 284 &  569 &   614 &  10 &  10 &  18 &   3 &    -    & 79.51   \\ \hline
  BL-ResNext50 Default & 64   &  128  &    256  &     512  &     3  &     4  &     6  &     3  &    77.02  & 75.50 \\
   \bottomrule
\end{tabular}    
\end{center}
\label{tab:blr1k_B}
\end{table}


\section{Conclusions} \label{sec:conclusions}
We described a novel NAS method based on polyharmonic splines that can perform search directly on large scale, imbalanced target datasets. We demonstrated how most common operations in deep neural networks can be included as variables in the search space of a spline modeling the accuracy of a given architecture. 
The number of evaluations required during the search phase of our NAS approach is proportional to the number of operations in the search space, not in the number of possible values they each operation could have, making the approach tractable at large scale.
We demonstrate the effectiveness of our method on the ImageNet22K benchmark~\cite{imagenet_cvpr09}, achieving a state of the art top-1 accuracy of $40.03\%$ . This result paves the way to apply polyharmonic spline based NAS to other architectures and operations within networks, potentially also including hyperparameters in the search space.

\section{Acknowledgement}
This research used resources of the Oak Ridge Leadership Computing Facility, which is a DOE Office of Science User Facility supported under Contract DE-AC05-00OR22725. It also used resources of the IBM T.J. Watson Research Center Scaling Cluster (WSC).


%
%
\bibliographystyle{splncs04}
\bibliography{main}
\end{document}